\documentclass[12pt]{article}
\usepackage[utf8]{inputenc}
\usepackage{graphicx}
\usepackage{mathtools}
\usepackage{amsmath}
\usepackage{caption}
\usepackage{subcaption}
\usepackage{lipsum}
\usepackage{float}
\usepackage{booktabs, multirow}
\usepackage{bbold}
\usepackage{soul}

\usepackage[english]{babel}
\usepackage{authblk}

\usepackage{lipsum}

\newcommand\blfootnote[1]{%
  \begingroup
  \renewcommand\thefootnote{}\footnote{#1}%
  \addtocounter{footnote}{-1}%
  \endgroup
}

\usepackage[letterpaper,top=0.8in,bottom=0.8in,left=0.5in,right=0.5in]{geometry}

\usepackage[colorlinks=true, allcolors=blue]{hyperref}

\title{BioADAPT-MRC: Adversarial Learning-based Domain Adaptation Improves Biomedical Machine Reading Comprehension Task}

\author[1,2, *]{Maria Mahbub}
\author[2]{Sudarshan Srinivasan}
\author[2]{Edmon Begoli}
\author[1]{Gregory D Peterson}
\affil[1]{Department of Electrical Engineering and Computer Science, University of Tennessee, Knoxville, TN, U.S.}
\affil[2]{Cyber Resilience and Intelligence Division, Oak Ridge National Laboratory, Oak Ridge, TN, U.S.}

\affil[*]{Corresponding author: mmahbub@vols.utk.edu}
\date{}

\begin{document}
\maketitle

\abstract{Biomedical machine reading comprehension (biomedical-MRC) aims to comprehend complex biomedical narratives and assist healthcare professionals in retrieving information from them. The high performance of modern neural network-based MRC systems depends on high-quality, large-scale, human-annotated training datasets. In the biomedical domain, a crucial challenge in creating such datasets is the requirement for domain knowledge, inducing the scarcity of labeled data and the need for transfer learning from the labeled general-purpose (source) domain to the biomedical (target) domain. However, there is a discrepancy in marginal distributions between the general-purpose and biomedical domains due to the variances in topics. Therefore, direct-transferring of learned representations from a model trained on a general-purpose domain to the biomedical domain can hurt the model's performance. We present an adversarial learning-based domain adaptation framework for the biomedical machine reading comprehension task (BioADAPT-MRC), a neural network-based method to address the discrepancies in the marginal distributions between the general and biomedical domain datasets. BioADAPT-MRC relaxes the need for generating pseudo labels for training a well-performing biomedical-MRC model. We extensively evaluate the performance of BioADAPT-MRC by comparing it with the best existing methods on three widely used benchmark biomedical-MRC datasets -- BioASQ-7b, BioASQ-8b, and BioASQ-9b. Our results suggest that without using any synthetic or human-annotated data from the biomedical domain, BioADAPT-MRC can achieve state-of-the-art performance on these datasets. BioADAPT-MRC is freely available as an open-source project at \url{https://github.com/mmahbub/BioADAPT-MRC}.\\
\blfootnote{This is the Authors' Original Version of the article, which has been accepted for publication in Bioinformatics Published by Oxford University Press.
All rights reserved.}
}


\maketitle

\section{Introduction}
\label{sec:intro}


During the consultation phase of primary patient care, for every two patients, healthcare professionals raise at least one question \cite{del2014clinical}.
Even though they can successfully find answers to 78\% of the pursued questions, they never pursue half of their questions because of time constraints and the suspicion that helpful answers do not exist, notwithstanding the availability of ample evidence \cite{del2014clinical, bastian2010seventy}.
Additionally, searching existing resources for reliable, relevant, and high-quality information poses an inconvenience for the clinicians on account of time limitation.
This phenomenon elicits the dependency on general-information electronic resources that are simple to use, such as Google \cite{hider2009information}.
Apart from the healthcare professionals, there is also a growing public interest in learning about their medical conditions online \cite{fox2013health}.
Nevertheless, the criteria for ranking search results by general-purpose search engines does not conform directly to the fundamentals of evidence-based medicine (EBM) and thus lacks rigor, reliability, and quality \cite{hider2009information}.

While traditional information retrieval (IR) systems somewhat mitigate this issue, it still requires four hours for a healthcare-information professional to find answers to queries related to complex biomedical resources \cite{russell2017expert}.
Compared to the IR systems that usually provide the users (general population or healthcare professionals) a group of documents to interpret and find the exact answers, biomedical machine reading comprehension (biomedical-MRC) systems can provide exact answers to user inquiries, saving both time and effort.

Machine reading comprehension (MRC) is a challenging task in natural language processing (NLP), aiming to teach and evaluate the machines to understand user-defined questions, read and comprehend input contexts (namely, context) and return answers from them.
The datasets in the MRC task consist of context-question-answer triplets where the question-answer pairs are considered labels.
With the development and availability of efficient computing hardware resources, researchers have developed several state-of-the-art (SOTA) neural network-based MRC systems capable of achieving analogous or superior to human-level performance on several benchmark MRC datasets \cite{devlin2018bert, rajpurkar-etal-2016-squad, joshi-etal-2017-triviaqa}.
However, this achievement is highly dependent on a large amount of high-quality human-annotated datasets that are used to train these systems \cite{rajpurkar-etal-2016-squad}.
For domain-specific MRC tasks, especially biomedical-MRC, building a high-quality labeled dataset, specifically, the question-answer pairs residing in the dataset requires undeniable effort and knowledge of subject matter experts.
This requirement leads to smaller biomedical-MRC datasets and, consequently, unreliably poor performance on the MRC task itself \cite{pergola2021boosting}.
Hence, developing an approach that can effectively leverage unlabeled or small-scale labeled datasets in training the biomedical-MRC model is crucial for improving performance.

Researchers have addressed this issue by using transfer learning, a learning process to transfer knowledge from a source domain to a target domain \cite{pan2009survey}.
In domain-specific MRC problems such as biomedical-MRC, the source domain is usually a general-purpose domain where a large-scale human-annotated MRC dataset is available.
The target domain, in this case, is the biomedical domain.
In this work, we focus on transferring the knowledge from an MRC model trained on a labeled general-purpose-domain dataset to the biomedical domain where only \textit{unlabeled contexts} are available.
Unlabeled contexts refer to only contexts in the MRC dataset with no question-answer pairs.

Often, directly transferring the knowledge representations (learned by an MRC model) from the source to the target domain can hurt the performance of the model because of the distributional discrepancies between the data seen at train and test time \cite{ganin2015unsupervised}.
Domain adaptation, a sub-setting of transfer learning \cite{pan2009survey}, aims at mitigating these discrepancies through \textit{simultaneous generation of feature representations that are discriminative from the viewpoint of the MRC task in the source domain and indiscriminative from the perspective of the shift in the marginal distributions between the source and target domains} \cite{ganin2015unsupervised}.

We propose \textbf{A}dversarial learning-based \textbf{D}omain ad\textbf{APT}ation framework for \textbf{Bio}-medical \textbf{M}achine \textbf{R}eading \textbf{C}omprehension (BioADAPT-MRC), a new framework that uses adversarial learning to generate domain-invariant feature representations for better domain adaptation in biomedical-MRC models.
In an adversarial learning framework, we train two adversaries (feature generator and discriminator) alternately or jointly against one another to generate domain-invariant features.
Domain-invariant features imply that the feature representations extracted from the source- and the target-domain samples are closer in the embedding space.

While other recent domain adaptation approaches for the MRC task focus on generating pseudo question-answer pairs to augment the training data \cite{golub-etal-2017-two, wang-etal-2019-adversarial}, we utilize only the unlabeled contexts from the target domain. This property makes our framework more suitable in cases where not only human-annotated dataset is scarce but also the generation of synthetic question-answer pairs is computationally expensive, and needs further validation from domain-experts (due to the sensitivity to the correctness of the domain knowledge).


We validate our proposed framework on three widely used benchmark datasets from the cornerstone challenge on biomedical question answering and semantic indexing, BioASQ \cite{tsatsaronis2015overview}, using their recommended evaluation metrics.
We empirically demonstrate that with the presence of no labeled data from the biomedical domain -- synthetic or human-annotated -- our framework can achieve SOTA performance on these datasets.
We further evaluate the domain adaptation capability of our framework by using clustering and dimensionality reduction techniques.
Additionally, we extend our framework to a semi-supervised setting
and use varying ratios of labeled target-domain data for evaluation.
Last but not least, we perform a thorough error analysis of our proposed framework to demonstrate its strengths and weaknesses.

The primary contributions of the paper are as follows:
(i) We propose BioADAPT-MRC, an adversarial learning-based domain adaptation framework that incorporates a domain similarity discriminator with an auxiliary task layer and aims at reducing the domain shift between high-resource general-purpose domain and low-resource biomedical domain.
(ii) We leverage the unlabeled contexts from the biomedical domain and thus relax the need for synthetic or human-annotated labels for target-domain data.
(iii) We further extend the learning paradigm of BioADAPT-MRC to a semi-supervised setting. We show that our framework can be successfully employed to improve the performance of a pre-trained language model in the presence of varying ratios of labeled target-domain data.
(iv) Through comprehensive evaluations and analyses on several benchmark datasets, we demonstrate the effectiveness of our proposed framework and its domain adaptation capability for biomedical-MRC.

\section{Background and related work}
\label{sec:back-rel}
In this paper, we focus on the biomedical-MRC task using the adversarial learning-based domain adaptation technique.
Thus, our work is in the confluence of two main research areas: biomedical machine reading comprehension and domain adaptation using adversarial learning.

\paragraph{\textbf{Biomedical machine reading comprehension:}}
In the biomedical machine reading comprehension (biomedical-MRC) task, the goal is to extract an answer span, given a user-defined question and a biomedical context.
In neural network-based (NN-based) biomedical-MRC systems, the question-context pairs are converted from discrete textual form to continuous high-dimensional vector form using word-embedding algorithms, such as word2vec \cite{mikolov2013efficient}, GloVe \cite{pennington2014glove}, FastText \cite{bojanowski2017enriching}, Bidirectional Encoder Representations from Transformers (BERT) \cite{devlin2018bert}, etc.
Among numerous architectural varieties of these NN-based MRC systems, the transformer-based pre-trained language models (PLMs) such as BERT are the current SOTA \cite{gu2021domain}.
The original BERT model is pre-trained on general-purpose English corpora.
Considering the semantic and syntactic uniqueness of the biomedical text, researchers have developed different variants of BERT models for the biomedical domain that are pre-trained on several biomedical corpora such as PMC full articles, PubMed abstracts, and  MIMIC datasets.
Some examples of such PLMs are BioBERT \cite{lee2020biobert}, PubMedBERT \cite{gu2021domain}, BioElectra \cite{raj2021bioelectra} which reportedly outperform the original BERT model in various biomedical NLP tasks.
These PLMs are used as trainable encoding modules (encoders) in downstream biomedical NLP tasks, such as biomedical named entity recognition \cite{naseem2021bioalbert}, clinical-note classification \cite{agnikula2021identification}, machine reading comprehension \cite{jeong2020transferability}, etc.
Usually, to accomplish the downstream tasks such as biomedical-MRC by transferring the knowledge from the PLMs, researchers add a few task-specific layers, commonly feed-forward neural network layers, at the end of the encoders \cite{hosein2019measuring, lee2020biobert, jeong2020transferability}.

\paragraph{\textbf{Transfer learning}:}
Transfer learning is an approach to transfer knowledge representations acquired from a widely explored domain/task (source), to a new or less explored domain/task (target) \cite{pan2009survey}. 
Adopting the notations provided by \cite{pan2009survey}, a domain $\mathcal{D}$ consists of a feature space $\mathcal{X}$ (different from the feature representation learned by the network) and a marginal distribution of the learning samples $X$ $\in \mathcal{X}$, $p(X)$.
In NLP, the marginal distributions are different when the languages are the same but the topics are different in the source and target domains \cite{bashath2022data}.
Considering a label space $\mathcal{Y}$, for a given domain $\mathcal{D}$, a task $\mathcal{T}$ can be described as $(\mathcal{Y}, \mathcal{F}(.))$, where $\mathcal{F}(.)$ is a predictor function learned from the training data $(X\in \mathcal{X}, Y\in \mathcal{Y})$.

Data scarcity in the target domain is often an impediment to the model's performance while training for a target task.
Transfer learning tackles this issue by aiming to improve generalizability on a target task using acquired knowledge from a source domain $\mathcal{D}_s$, a target domain $\mathcal{D}_t$ along with their respective associated tasks $\mathcal{T}_s$ and $\mathcal{T}_t$.
For this work, we use \textit{transductive transfer learning} which uses labeled source domain, unlabeled target domain, identical source and target tasks, and different source and target domains.
Depending on the similarity in the feature spaces, there are two cases of transductive transfer learning: i) different feature spaces for the source and target domains, e.g., cross-lingual transfer learning, ii) identical feature spaces, but different marginal probability distributions for the source-domain and target-domain samples, e.g., transfer learning between two domains with the same language but different topics \cite{bashath2022data}.
In this paper, we use the latter case of transductive transfer learning to train the biomedical-MRC system, otherwise known as domain adaptation \cite{pan2009survey}.

\paragraph{\textbf{Domain adaptation:}}
Domain adaptation aims at increasing the generalizability of machine learning models when posed with unlabeled or very few labeled data from the target domain by generating domain invariant representation \cite{glorot2011domain}.
One can enforce the learning of domain-invariant features in machine learning models by implementing the adversarial learning framework \cite{ganin2015unsupervised, tzeng2017adversarial}.
In the adversarial setting, usually, a domain discriminator is incorporated into the MRC framework where besides performing the MRC task, the goal is to attempt at fooling the discriminator by generating domain invariant features \cite{wang-etal-2019-adversarial}.
Researchers have successfully applied domain adaptation in many tasks such as sentiment classification \cite{glorot2011domain}, speech recognition \cite{sun2017unsupervised}, neural machine translation \cite{thompson2019overcoming}, named entity recognition \cite{vu2020effective}, and image segmentation \cite{guan2021scale}.
However, compared to these tasks, the application of domain adaptation to the MRC task poses one additional challenge apart from missing answers -- the missing questions in the target domain.
Over recent years, researchers have proposed various methods to generate synthetic question-answer pairs from unlabeled contexts.
For example, \cite{wang-etal-2019-adversarial} used NER and Bi-LSTM, \cite{golub-etal-2017-two} used conditional probability, IOB tagger, and Bi-LSTM, and \cite{Yue2021CliniQG4QAGD} used seq2seq model with an attention mechanism to generate pseudo question-answer pairs.
A multi-task learning approach has also been used for domain adaptation in MRC tasks \cite{Nishida2020UnsupervisedDA}.

Among these works in MRC and domain adaptation, the AdaMRC model proposed by \cite{wang-etal-2019-adversarial} focuses on learning domain-invariant features in an adversarial setting as ours.
However, the main differences between BioADAPT-MRC and AdaMRC are as follows:
(i) While AdaMRC uses synthetically generated question-answer pairs to augment the target-domain dataset, BioADAPT-MRC directly uses the unlabeled contexts and thus relaxes the need for synthetic question-answer pairs.
In later sections, we show that although synthetic questions can improve the performance of MRC tasks for various target domains such as Wikipedia, web search log, and news \cite{wang-etal-2019-adversarial}, they can hurt the performance of the MRC task for the biomedical domain.
(ii) While AdaMRC uses the binary classification loss, BioADAPT-MRC uses triplet loss to minimize the domain shift between the source and target domains.
Unlike binary classification loss, triplet loss considers both similarity and dissimilarity between two samples for gradient update and is known to be successful in deep metric learning where the aim is to map semantically similar instances closer in the embedding space and vice versa \cite{kim2018attention, chen2018darkrank, kaya2019deep}.
Moreover, triplet loss makes BioADAPT-MRC directly applicable to domain adaptation among more than two domains.
While multiple prior works in computer vision have successfully used triplet loss for domain adaptation in numerous applications \cite{laiz2019using, wen2018improving}, to the best of our knowledge, ours is the first in the application of the MRC task in NLP.
(iii) To improve performance and stabilize the training process in the adversarial domain adaptation framework, BioADAPT-MRC uses an auxiliary task layer, similar to AC-GAN \cite{odena2017conditional}.

\section{Materials and methods}
\label{sec:mat-meth}
In this section, we discuss our adversarial learning-based domain adaptation framework for the biomedical-MRC task.

\begin{figure}[!htpb]
\centerline{\includegraphics[width=1.\textwidth]{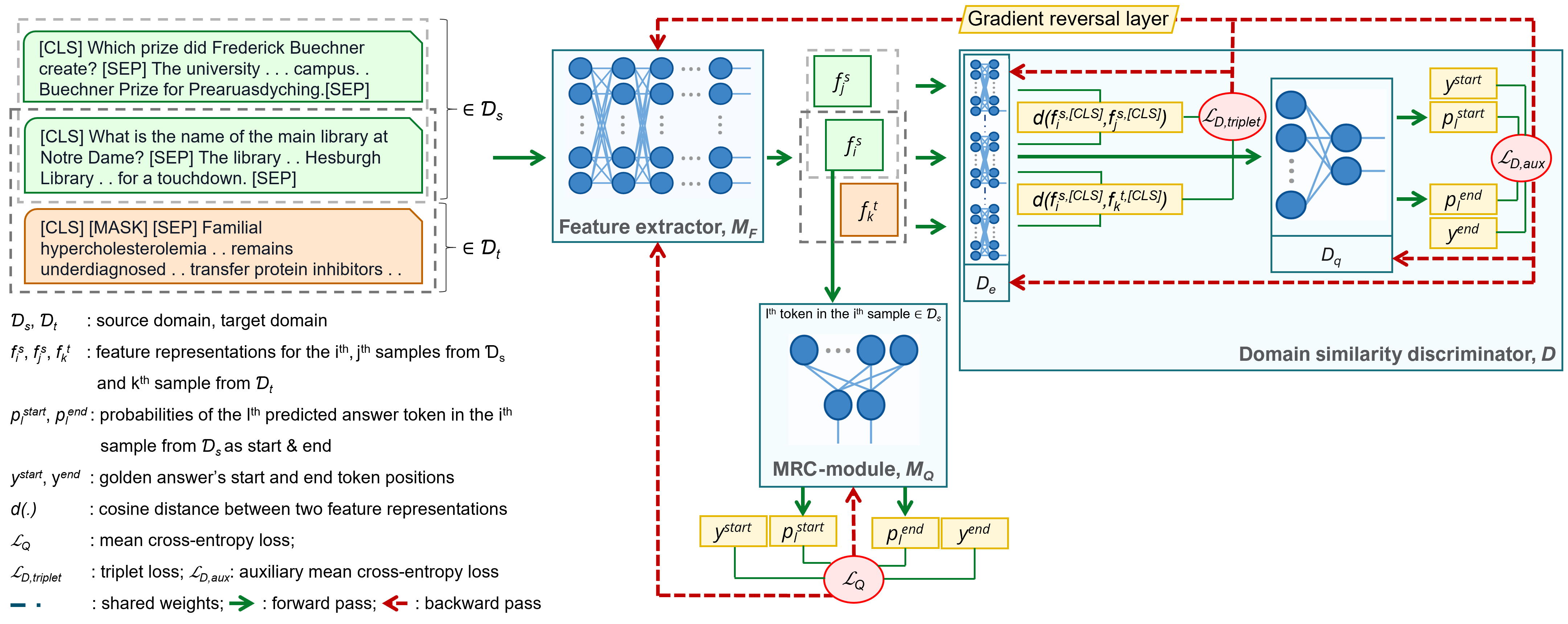}}
\caption{BioADAPT-MRC: an adversarial learning-based domain adaptation framework for biomedical machine reading comprehension task.
The framework has three main components: (i) feature extractor $M_F$, (ii) MRC-module $M_Q$, and (iii) domain similarity discriminator $D$.}
\label{fig:framework}
\end{figure}
\subsection{Problem definition}

Given an unlabeled target domain $\mathcal{D}_t$ and a labeled source domain $\mathcal{D}_s$ along with their respective learning tasks $\mathcal{T}_t$ and $\mathcal{T}_s$, we assume that $\mathcal{T}_t = \mathcal{T}_s$ and $\mathcal{D}_t \ne \mathcal{D}_s$ because of $p(X_t\in \mathcal{X}) \ne p(X_s\in \mathcal{X})$, where $p(.)$ is the marginal probability distribution, $X_t$ and $X_s$ are learning samples from the target and source domains, respectively.
Thus, while the tasks are identical, the domains are different due to different marginal probability distributions in their data.

In this work, $\mathcal{D}_t$ is the biomedical domain where only unlabeled biomedical contexts are available, and $\mathcal{D}_s$ is the general-purpose domain where large-scale labeled data are available.
As mentioned in Section \ref{sec:back-rel}, despite having the same language, differences in the topics between two domains cause the domains to be different because of the dissimilarities in $p(.)$.
In this work, we consider that the general-purpose and the biomedical domains have different topics.
Thus, we assume that $p(X_t) \ne p(X_s)$.


The task for both domains is extractive MRC.
Given a question $Q = \{q_1, q_2, ..., q_n\}$ and a context $C = \{c_1, c_2, ..., c_m\}$, extractive MRC predicts the start and end positions $a^{start}$ and $a^{end}$, respectively, of the answer span $A = \{c_i\}_{a^{start}}^{a^{end}}$ in $C$ such that there exists one and only one answer span consisting of continuous tokens in the context.
Here, $q_i$ denotes the $i^{th}$ token in the question, $c_i$ denotes the $i^{th}$ token in the context, and $n$, $m$ respectively denote the number of tokens in $Q$ and $C$.


\subsection{BioADAPT-MRC}

Given the labeled and unlabeled inputs respectively from the source and target domains, our proposed framework BioADAPT-MRC aims at achieving the following two objectives:
i) predicting the answer spans from the provided contexts,
ii) addressing the discrepancies in the marginal distributions between the inputs in the source and target domains by generating domain-invariant features.
Figure~\ref{fig:framework} demonstrates the three primary components of the BioADAPT-MRC framework:

\begin{itemize}
    \item \textit{Feature extractor} accepts a text sequence and encodes it into a high-dimensional continuous vector representation.
    
    \item \textit{MRC-module} accepts the encoded representation from either the source domain (training time) or the target domain (test time), then predicts the start and end positions of the answer span $A$ in $C$.
    
    \item \textit{Domain similarity discriminator} accepts the encoded representations from the source and target domains and learns to distinguish between them.
\end{itemize}

\subsubsection{Feature extractor}
\label{sec:mat-fe}

Given an input sample from either domain, the feature extractor $M_F(.)$ maps it to a common feature space $F$:
\begin{equation}
    f_i = M_F(X_i), X_i\in\ \mathcal{D}_s \cup \mathcal{D}_t\
\end{equation}
Here, $f_i$ is the extracted feature for the $i^{th}$ input sample $X_i$ from either $\mathcal{D}_s$ or $\mathcal{D}_t$.
We utilize the encoder of the PLM BioELECTRA \cite{raj2021bioelectra} as the feature extractor.
We choose BioELECTRA for the following reason:
While biomedical domain-specific BERT models such as BioBERT, SciBERT outperform the original BERT models in several biomedical NLP tasks \cite{alsentzer2019publicly}, BioELECTRA has the best performance scores on the \textit{Biomedical Language Understanding and Reasoning Benchmark (BLURB)} \cite{gu2021domain}.

As mentioned in Section \ref{sec:back-rel}, the features in this task are trainable, high-dimensional word embeddings extracted from the question-context pairs.
To generate these word embeddings, the BioELECTRA model utilizes the transformer-based architecture from one of the BERT-variants, ELECTRA.
The ELECTRA model has 12 layers, 768 hidden size, 3072 feed-forward network (FFN) inner hidden size, and 12 attention heads per layer \cite{clark2020electra}.
The pre-training corpora for BioELECTRA is 3.2 million PubMed Central full-text articles and 22 million PubMed abstracts, and the pre-training task is the replaced token prediction task.
BioELECTRA has a vocabulary of size 30522.
The maximum number of tokens per input can be 512, where the embedding dimension of each token is 768.
For each pair, the final tokenized input of the BioELECTRA model is $\{[CLS], Q, [SEP], C, [SEP]\}$.
Here, $Q, C$ are respectively the tokens from the question and the context, $[CLS]$ is a special token that can be considered to have an accumulated representation of the input sequence \cite{devlin2018bert} and used for classification tasks, $[SEP]$ is another special token that separates two consecutive sequences.
Note that, since the samples in the target domain are unlabeled, in place of the question tokens $Q$, we use a special token $[MASK]$ to maintain consistency in the structure of the tokenized samples.

\subsubsection{MRC-module}
\label{sec:mat-mrc}

As the MRC-module $M_Q(.)$, we add a simple fully-connected layer with hidden size $H=768$ on top of the feature extractor $M_F(.)$ and use the softmax activation function to generate probability distributions for start and end token positions following Equation \ref{eq:softmax}.
\begin{equation} \label{eq:softmax}
    p_l^{start} = \frac{exp^(W^{start} \cdot h_l)}{\sum_{k=1}^{n_{seq}} exp^{W^{start} \cdot h_k}};~~
    p_l^{end} = \frac{exp^(W^{end} \cdot h_l)}{\sum_{k=1}^{n_{seq}} exp^{W^{end} \cdot h_k}}
\end{equation}
Here, $p_l^{start}$ and $p_l^{end}$ are the probabilities of the $l^{th}$ token to be predicted as $start$ and $end$ respectively, $h_l \in \mathbb{R}^H$ is the hidden representation vector of the $l^{th}$ token, 
$W^{start}\in \mathbb{R}^{H}$ and $W^{end}\in \mathbb{R}^{H}$ are two trainable weight matrices,
$n_{seq}$ is the input sequence length.
We use the cross-entropy loss $\mathcal{L}_Q$ on the predicted answer positions as the objective function for the  $M_Q$.
Since for each answer span prediction, we get two predicted outputs for the start and end positions, we average the total cross-entropy loss as shown in Equation \ref{eq:ce_loss}.
\begin{equation}
\label{eq:ce_loss}
    \mathcal{L}_Q = -\frac{1}{2} (log p_{y^{start}}^{start} + log p_{y^{end}}^{end})
\end{equation}
Here the golden answer’s start and end token positions are represented by $y^{start}$ and $y^{end}$, respectively.
During test phase, the predicted answer span is selected based on the positions of the highest probabilities from $p_{k\in [1,n_{seq}]}^{start}$ and $p_{k\in [1,n_{seq}]}^{end}$.

\subsubsection{Domain similarity discriminator}
\label{sec:dom-sim-disc}
The domain similarity discriminator $D(.)$ addresses the domain variance between two domains (caused by the discrepancies in the marginal probability distributions), as follows:
In the adversarial setting, $D(.)$ learns to distinguish between the feature representations of the source-domain and target-domain samples generated by the feature extractor.
$D(.)$ then penalizes the feature extractor for producing domain-variant feature representations and thus promotes the generation of domain-invariant features.
$D(.)$ uses cosine distance between the feature representations of the input samples to distinguish between the domains.
We consider that two samples are closer in the embedding space and thus have a greater chance to be in the same domain if their feature representations have a smaller cosine distance between them and vice-versa. 
The input of $D(.)$ is a triplet $(f_k^t, f_i^s, f_j^s)$ where $f_k^t$, $f_i^s$, and $f_j^s$ are respectively the feature representations of the $k^{th}$ sample $X_k^t$ from the target domain and $i^{th}$ sample $X_i^s$ and $j^{th}$ sample $X_j^s$ from the source domain extracted by $M_F(.)$.
The triplet is then split into two distinct pairs, consisting of $(f_i^s, f_k^t)$ and $(f_i^s, f_j^s)$.
As indicated in Figure \ref{fig:framework}, upon receiving each triplet, $D$ accomplishes two tasks:
i) measures the similarity between $(f_i^s, f_j^s)$ and dissimilarity between $(f_i^s, f_k^t)$,
ii) performs MRC task similar to $M_Q$ for the source sample $X_i^s$.

For the first task, we introduce a Siamese network \cite{bromley1993signature} $D_e$ with a single transformer encoder layer.
$D_e$ acts as a function that helps estimate the similarity and dissimilarity between the received pairs.
Considering the success of the $BERT$ models in many NLP tasks, for the Siamese network, we adopt the same architecture as any encoder layer in the $BERT_{Base}$ model which has 12 attention heads, 768 embedding dimensions, 3072 FFN inner hidden size with 10\% dropout rate and `GeLU' activation function.
We encode the input pairs $(f_i^s, f_k^t)$ and $(f_i^s, f_j^s)$, using the same encoder network $D_e$.
Considering the role of the special token $[CLS]$ as explained in \ref{sec:mat-fe}, to let $D(.)$ differentiate whether the pairs are from the same domain or not, we extract the $[CLS]$ token representations from $D_e$ for $f_k^t, f_i^s,$ and $f_j^s$.
We then use these $[CLS]$ token representations to calculate the domain-similarity and dissimilarity via triplet loss function \cite{weinberger2005distance} and use it as the learning objective of the discriminator $D(.)$ as shown in Equation \ref{eq:triplet}.

\begin{equation}
\label{eq:triplet}
\begin{aligned}
    \mathcal{L}_{D,triplet} = max\{d(f_i^{s, [CLS]}, f_j^{s, [CLS]}) \\- d(f_i^{s, [CLS]}, f_k^{t, [CLS]}) + \alpha, 0\}
\end{aligned}
\end{equation}
Here, $f_i^{s, [CLS]}, f_j^{s, [CLS]}$, $f_k^{t, [CLS]}$ are respectively the $[CLS]$ token representations of $i^{th}$ and $j^{th}$ samples from the source domain and the $k^{th}$ sample from the target domain where $i \ne j$, $d(.) = 1.0 - CosSim(.)$ is the cosine distance where $CosSim(.)$ is the cosine similarity, $\alpha$ is the non-negative margin representing the minimum difference between $d(f_i^{s, [CLS]}, f_j^{s, [CLS]})$ and $d(f_i^{s, [CLS]}, f_k^{t, [CLS]})$ that is required for the triplet loss to be $0$.
To optimize $D(.)$, $\mathcal{L}_{D, triplet}$ aims at minimizing $d(.)$ between the samples from the source domain and maximizing $d(.)$ between the samples from the source and target domains.
Using triplet loss, our discriminator efficiently employs both similar and dissimilar information extracted by the feature extractor component of the model.

While using triplet loss in the adversarial setting, we also consider that the triplet loss might make representations from the source domain dissimilar.
Therefore, as an additional experiment, we try optimizing the discriminator by minimizing a distance-based loss, which is equivalent to just minimizing the distance $d(.)$ between the source- and the target-domain sample representations in the adversarial setting.
We demonstrate the comparison between these two approaches in Section \ref{sec:res-loss}.


Adversarial learning for domain adaptation is known to be unstable \cite{wulfmeier2017addressing, rios2018generalizing}.
To stabilize the training process, we use the concept of AC-GAN \cite{odena2017conditional}.
AC-GAN uses an auxiliary task layer on top of the discriminator and appears to stabilize the adversarial learning procedure and improve the performance \cite{odena2017conditional}.

Following this concept, for the second task of $D(.)$, we introduce an MRC-module $D_q$ similar to $M_Q$ on top of $D_e$ as an auxiliary task layer.
$D_q$ enforces that the discriminator does not lose task-specific information while learning to encode domain-variant features.
Later in Sections \ref{sec:res-exp} and \ref{sec:res-stab}, we demonstrate the effectiveness of the auxiliary layer by performing an ablation study and a stability analysis.
The input of $D_q$ is the output of $D_e$ for $f_i^s$ and the output is the probability distributions for the start and end token positions of the answer span, similar to $(p_i^{start}, p_j^{end})$.
Thus, the loss function $\mathcal{L}_{D,aux}$ for $D_q$ is the same as $\mathcal{L}_Q$.
The final loss function $\mathcal{L}_D$ for $D(.)$ is shown in Equation \ref{eq:disc_loss}.
\begin{equation} \label{eq:disc_loss}
    \mathcal{L}_D = \mathcal{L}_{D,triplet} + \mathcal{L}_{D,aux}
\end{equation}

\subsubsection{Cost function}
To eliminate domain shift by learning domain-invariant features, we integrate $M_F(.)$, $M_Q(.)$, and $D(.)$ into adversarial learning framework where we update $M_F$ and $M_Q$ to maximize $\mathcal{L}_D$ and minimize $\mathcal{L}_Q$ while updating $D$ to minimize $\mathcal{L}_D$.
Thus, the cost function $\mathcal{L}_{total}$ of the BioADAPT-MRC framework consists of $\mathcal{L}_Q$ and $\mathcal{L}_D$ as shown in Equation \ref{eq:total_loss} and is optimized end-to-end:
\begin{equation}\label{eq:total_loss}
    \mathcal{L}_{total} = \mathcal{L}_Q - \lambda \mathcal{L}_D
\end{equation}
Here, $\lambda$ is a regularization parameter to balance $\mathcal{L}_Q$ and $\mathcal{L}_D$.
Unlike the original adversarial learning framework proposed in GAN, where the adversaries are updated alternately \cite{goodfellow2014generative}, we perform joint optimization for all three components of our model using the gradient-reversal layer \cite{ganin2015unsupervised}, as suggested by \cite{chen2018adversarial}.


\section{Results and discussion}
\label{sec:res-disc}

We perform an extensive study to evaluate the proposed framework and compare with the SOTA biomedical-MRC methods on a collection of publicly available and widely used benchmark biomedical-MRC datasets.

\subsection{Dataset}

To demonstrate the effectiveness of our framework, we evaluate BioADAPT-MRC and compare it with the SOTA methods on three biomedical-MRC datasets from the BioASQ annual challenge  \cite{tsatsaronis2015overview}.
The BioASQ competition has been organized since 2013 and consists of two large-scale biomedical NLP tasks, one of which is question-answering (task B).
Among four types of questions in task B, the factoid questions resemble the extractive biomedical-MRC task.
As such, we utilize only the factoid MRC data from the BioASQ challenges held in 2019 (BioASQ-7b), 2020 (BioASQ-8b), and 2021 (BioASQ-9b) as the target-domain datasets to verify our model.
These datasets were created from the search engine for biomedical literature, PubMed, with the help of domain experts.
Note that, for training, our framework requires only unlabeled contexts in the target domain.
As such, we only consider the contexts in the BioASQ-7b,8b, and 9b training sets and disregard the question-answer pairs.
The details on the availability of the training data and the pre-processing steps are provided below:

We reinforce the reliability of the comparison of experimental results with other methods and validate the efficacy of the BioADAPT-MRC framework by using the pre-processed training set for BioASQ-7b and 8b, provided by \cite{yoon2019pre} and \cite{jeong2020transferability}, respectively.
For BioASQ-9b, we pre-process the training data by retrieving full abstracts from PubMed using their provided PMIDs.
We use these retrieved abstracts as the contexts in the BioASQ-9b training set.
Since our framework requires no label (question-answer pair) in the training set of the target domain, for all BioASQ training sets, we disregard each question by assigning a $[MASK]$ token and each answer by replacing by an empty string.
At test time, we use the golden enriched test sets -- BioASQ-7b, 8b, and 9b -- from the BioASQ challenges.

\begin{table}[!htb]
\centering
\caption{Statistics of the datasets used in the experiments.\label{Tab:data_stat}}
\resizebox{1\textwidth}{!}{%
{\begin{tabular}{@{}lllll@{}}\toprule
Dataset   & Training set  & Training set    & Target to source  & Test set \\ 
name      & (raw)         & (pre-processed) & ratio in training set &          \\   
\midrule
SQuAD-1.1      & 87,599            & 87,599                      & – & –        \\
BioASQ-7b      & 779               & 5,537                       & $\sim$1:16 & 162      \\
BioASQ-8b      & 941               & 10,147                      & $\sim$1:9  & 151      \\
BioASQ-9b      & 1,092             & 13,178                      & $\sim$1:7  & 163      \\
\hline
\end{tabular}}
}
\end{table}


Similar to the previous studies \cite{yoon2019pre, jeong2020transferability}, as the source-domain dataset, we use SQuAD-1.1 \cite{rajpurkar-etal-2016-squad}, which was developed from Wikipedia articles by crowd-workers.
Table \ref{Tab:data_stat} shows the basic statistical information of all datasets used in the experiments.
As shown, the number of training data samples in the source domain is noticeably higher than that of the target domain.
The details on experimental setup and training configurations are provided below:

The BioADAPT-MRC framework consists of three main components: feature extractor, MRC-module, and discriminator.
In the implementation of the framework with PyTorch \cite{paszke2019pytorch}, we initialize the feature extractor with the parameters from the pre-trained BioELECTRA model using the huggingface API \cite{wolf2019huggingface}.
For the parameters in the MRC-module and the discriminator, we perform random initialization.
For tokenization, we set the maximum query length to 64, the maximum answer length to 30, the maximum sequence length to 384, and the document stride to 128, as suggested in \cite{raj2021bioelectra, devlin2018bert, lee2020biobert}.
For optimizing the hyperparameters -- learning rate, batch size, and the regularization parameter $\lambda$, we use a set of randomly selected 5000 samples from the SQuAD development dataset.
We then empirically determined that learning rate 5e-5 and batch size 35 are the best choices for our experiments and for the hardware resources we are using.
We also empirically set the regularization parameter $\lambda$ to 0 and then increase it by 0.01  every 10 epochs up to 0.04.

For each step in the training epoch, we randomly select two samples from the source domain dataset SQuAD and one sample from the target domain dataset BioASQ-7b/8b/9b.
We run all our experiments on a Linux server with Intel(R) Xeon(R) Gold 6130 CPU @ 2.10GHz and a single Tesla V100-SXM2-16GB GPU.

\subsection{Metrics}

For evaluation, we use three metrics used in the MRC task in the official BioASQ challenge: strict accuracy (SAcc), lenient accuracy (LAcc), and mean reciprocal rank (MRR).
The BioASQ challenge requires the participant systems to predict the five best-matched answer spans extracted from the context(s) in a decreasing order based on confidence score.
In the test set, for each question, the biomedical experts in the BioASQ team provided one golden answer extracted from the context.
Both golden answers and predicted answer spans are used to calculate the SAcc, LAcc, and MRR scores, as shown in Equation \ref{eq:sacc}.
SAcc shows the models' capability to find exact answer location, LAcc determines the models' understanding of predicted answers' range and MRR reflects the quality of the predicted answer spans \cite{tsatsaronis2015overview}:
\begin{equation}
\label{eq:sacc}
    SAcc = \frac{c_1}{n_{test}};~~LAcc = \frac{c_5}{n_{test}}; ~~MRR = \frac{1}{n_{test}}\sum_{i=1}^{n_{test}}\frac{1}{r(i)}
\end{equation}
Here, $c_1$ is the number of questions correctly answered by the predicted answer span with the highest confidence score, $c_5$ is the number of questions answered correctly by any of the five predicted answer spans, $n_{test}$ is the number of questions in the test set, and $r(i)$ is the rank of the golden answer among all five predicted answer spans for the $i^{th}$ question.
If the golden answer does not belong to the five predicted answer spans, we consider $\frac{1}{r(i)}=0$. 
We implement these metrics by leveraging the publicly available tools provided by the official BioASQ challenge at https://github.com/BioASQ/Evaluation-Measures.
\begin{table}[!htbp]
\caption{Performance of BioADAPT-MRC compared with the best scores on BioASQ-7b, BioASQ-8b, and BioASQ-9b test sets. The best and the second-best scores are respectively highlighted in bold and italic. `–' indicates that the corresponding source did not report the scores. * denotes previously best-performing method for BioASQ-7B and BioASQ-8B.
\label{Tab:com_exp}}
\resizebox{1\textwidth}{!}{%
{\begin{tabular}{@{}l lll lll lll@{}}\toprule
\multirow{2}{*}{Model}         & \multicolumn{3}{c}{BioASQ-7b} & \multicolumn{3}{c}{BioASQ-8b} & \multicolumn{3}{c}{BioASQ-9b}\\
                                 \cmidrule(lr){2-4}              \cmidrule(lr){5-7}              \cmidrule(lr){8-10}
                                                                    & SAcc     & LAcc     & MRR     & SAcc     & LAcc     & MRR       & SAcc     & LAcc     & MRR  \\ \midrule
Google \cite{hosein2019measuring}                              & 0.4201   & 0.5822   & 0.4798  & –        & –        & –       & –        & –        & –       \\
BioBERT \cite{yoon2019pre}                                     & 0.4367   & 0.6274   & 0.5115  & –        & –        & –       & –        & –        & –       \\
UNCC \cite{telukuntla2019uncc}                                 & 0.3554   & 0.4922   & 0.4063  & –        & –        & –       & –        & –        & –       \\
Umass \cite{kommaraju2020unsupervised}                         & –        & –        & –       & 0.3133   & 0.4798   & 0.3780  & –        & –        & –       \\
KU-DMIS-2020 \cite{jeong2020transferability}                   & \textbf{0.4510}   & 0.6245   & 0.5163  & 0.3819   & 0.5719   & 0.4593  & –        & –        & –       \\
BioQAExternalFeatures \cite{xu2021external}*               & 0.4444   & \textit{0.6419}   & \textit{0.5165}  & \textbf{0.3937}   & \textit{0.6098}   & \textit{0.4688}  & –        & –        & –       \\
BioASQ-9b Challenge - Best system (Ir\_sys2)  & –        & –        & –       & –        & –        & –       & 0.5031   & 0.6626   & 0.5667  \\
BioASQ-9b Challenge - Hypothetical system & –        & –        & –       & –        & –        & –       & \textbf{0.5399}   & \textit{0.7300}   & \textit{0.6017}  \\
AdaMRC~\cite{wang-etal-2019-adversarial}       & 0.4321                   & 0.6235                   & 0.5136                  & 0.3510                   & 0.5828                   & 0.4455                  & \textit{0.5337}                   & 0.7117                   & 0.6001                  \\
BioADAPT-MRC                                                         & \textit{0.4506}   & \textbf{0.6420}   & \textbf{0.5289}  & \textit{0.3841}   & \textbf{0.6225}   & \textbf{0.4749}  & \textbf{0.5399}   & \textbf{0.7423}   & \textbf{0.6187}  \\
\hline
\end{tabular}}
}
\end{table}
\begin{table}[!t]
\caption{Test scores for ablation experiments of BioADAPT-MRC. The best and the second-best scores are respectively highlighted in bold and italic.
\label{Tab:abl_exp}}
\resizebox{1\textwidth}{!}{%
{\begin{tabular}{@{}l lll lll lll@{}}\toprule
\multirow{2}{*}{Model}                       & \multicolumn{3}{c}{BioASQ-7b} & \multicolumn{3}{c}{BioASQ-8b} & \multicolumn{3}{c}{BioASQ-9b}\\
                                               \cmidrule(lr){2-4}              \cmidrule(lr){5-7}              \cmidrule(lr){8-10}
                                             & SAcc     & LAcc     & MRR     & SAcc     & LAcc     & MRR     & SAcc     & LAcc     & MRR  \\ \midrule
Baseline                                    & 0.4136                   & \textit{0.6296}          & 0.5056                  & 0.3642                   & 0.5960                   & 0.4602                  & 0.5092                   & 0.7362                   & 0.6010                  \\
BioADAPT-MRC (no auxiliary layer)  & \textit{0.4259}          & \textit{0.6296}          & \textit{0.5146}         & \textit{0.3775}          & \textit{0.6026}          & \textit{0.4679}         & \textit{0.5276}          & \textbf{0.7485}          & \textit{0.6142}         \\
BioADAPT-MRC  & \textbf{0.4506}          & \textbf{0.6420}          & \textbf{0.5289}         & \textbf{0.3841}          & \textbf{0.6225}          & \textbf{0.4749}         & \textbf{0.5399}          & \textit{0.7423}          & \textbf{0.6187} \\     
\hline
\end{tabular}}
}
\end{table}


\subsection{Method comparison}

We compare the test-time performance of BioADAPT-MRC on BioASQ-7b and 8b with six best performing models selected based on related published articles: Google \cite{hosein2019measuring}, BioBERT \cite{yoon2019pre}, UNCC \cite{telukuntla2019uncc}, Umass \cite{kommaraju2020unsupervised}, KU-DMIS-2020 \cite{jeong2020transferability}, and BioQAExternalFeatures \cite{xu2021external}.
For BioASQ-9b, we pick the best performing system Ir\_sys2 from the BioASQ-9b leaderboard (available at: \url{http://participants-area.bioasq.org/results/9b/phaseB/}).
\\
We also consider a hypothetical system that we would get for BioASQ-9b if that system would achieve the highest SAcc, LAcc, and MRR scores on the leaderboard in all five batches of this test set. Note that, in reality, no individual system in the competition achieved the highest scores for all three metrics in all the batches.

In addition to these models, we also compare the performance of BioADAPT-MRC with AdaMRC \cite{wang-etal-2019-adversarial}, a state-of-the-art domain adaptation method for the MRC task.
We provide brief descriptions of these aforementioned models below:

\begin{itemize}
    \item \textit{Google:} 
    Authors in \cite{hosein2019measuring} have used sequential transfer learning for biomedical-MRC task by fine-tuning a BERT PLM on two labeled general-purpose source-domain datasets -- NQ \cite{kwiatkowski2019natural} and CoQA \cite{reddy2019coqa}, and a labeled target-domain dataset -- BioASQ \cite{tsatsaronis2015overview}.
    \item \textit{BioBERT:} 
    Authors in \cite{yoon2019pre} have taken a similar approach as \cite{hosein2019measuring} but fine-tuned a BioBERT PLM and used SQuAD as the labeled general-purpose source-domain dataset.
    Their system secured first place in the BioASQ-7b challenge.
    \item \textit{UNCC:}
    In addition to the approach taken in \cite{yoon2019pre}, authors in \cite{telukuntla2019uncc} have used lexical answer types (LAT) as additional input features.
    \item \textit{Umass:}
    Authors in \cite{kommaraju2020unsupervised} have used labeled SQuAD data and unlabeled BioASQ data for transfer learning with BioBERT and SciBERT, along with a focus on a self-denoising task that helps learn mentions of biomedical named entities in the contexts.
    \item \textit{KU-DMIS-2020:}
    Authors in \cite{jeong2020transferability} have introduced the natural language inference (NLI) task in the pipeline of sequential transfer learning, where it fine-tunes BioBERT on labeled SQuAD, MNLI, and BioASQ datasets.
    \item \textit{BioQAExternalFeatures:} 
    The current SOTA method on BioASQ-7b and 8b was proposed by authors in \cite{xu2021external}.
    It uses externally extracted syntactic and lexical features in addition to the approach taken by authors in \cite{yoon2019pre}, possibly exposing to different adversarial attacks that may leverage syntactic and lexical knowledge-base from the dataset \cite{qi2021hidden}.
    \item \textit{BioASQ-9b Challenge - Best system (Ir\_sys2):} 
    For BioASQ-9b, we pick the best performing model \textit{Ir\_sys2} 
    from the BioASQ-9b leaderboard (available at \url{http://participants-area.bioasq.org/results/9b/phaseB/}).
    Each year the BioASQ question-answering challenge publishes five batches of the test set (e.g., Task 9b - Test batch 1, Test batch 2, etc.), and the participants submit their system evaluations for each of these batches separately.
    To find the best-performing system from the leaderboard for BioASQ-9b, we take the following steps:
    First, we discard the systems that did not report scores for all five batches.
    Then we calculate the weighted average of SAcc, LAcc, and MRR scores (across five batches) of the remaining systems, which is equivalent to reporting the results on the whole BioASQ-9b dataset (see Table \ref{tab:9b_lead}).
    We do this because the published articles on the compared state-of-the-art models (for BioASQ-7b and 8b) reported SAcc, LAcc, and MRR scores on the whole datasets.
    Finally, to select the best system for BioASQ-9b, we average these mean SAcc, mean LAcc, and mean MRR scores (shown in the last column of Table \ref{tab:9b_lead}) for each system and find that Ir\_sys2 is the best performing system.
    Ir\_sys2 achieved the highest SAcc and MRR scores.
    For the LAcc score, Ir\_sys2 achieved the second-best score with a small difference of 0.006.
    \item \textit{BioASQ-9b Challenge - Hypothetical system:} 
    As another compared baseline for BioASQ-9b, we also considered a hypothetical system with the best SAcc, LAcc, and MRR scores in all five batches of the BioASQ-9b test set.
    Note that no system on the leaderboard achieved the highest scores in all three metrics for all five batches in the test set.
    Later in Section \ref{sec:res-exp}, we show that BioADAPT-MRC outperforms even this hypothetical system.
    \item \textit{AdaMRC:}
    Authors in \cite{wang-etal-2019-adversarial} have used an adversarial domain adaptation framework that uses a binary domain classifier, a synthetic-question generator, and an MRC-module.
    As the MRC module, AdaMRC uses SAN, BiDAF, BERT, and ELMo.
    AdaMRC has shown to improve performance for target domains such as Wikipedia (SQuAD dataset), news (NewsQA dataset \cite{trischler-etal-2017-newsqa}), and web search log (MS MARCO dataset \cite{nguyen2016ms}).
\end{itemize}

\subsection{Experimental results}
\label{sec:res-exp}
Table \ref{Tab:com_exp} shows the comparison of BioADAPT-MRC with the state-of-the-art biomedical-MRC methods on BioASQ-7b, 8b, and 9b.
As shown, BioADAPT-MRC improves on both LAcc and MRR when tested on all three BioASQ test sets and achieves the best performance.
We also notice that while our model achieves the highest SAcc score for BioASQ-9b, it achieves the second-best SAcc scores for BioASQ-7b and 8b.
The higher SAcc and LAcc scores imply that our model is able to correctly extract complete answers from the given contexts more frequently than the previous methods.
The higher MRR scores, on the other hand, reflect our model's ability to extract complete answers with higher probability than the previous methods.
In contrast to the previous works, our method uses no label information (question-answer pairs) during the training process and has still been able to achieve good performance, implying the effectiveness of our proposed framework.

As explained in Section \ref{sec:mat-meth}, in the framework, we propose a domain similarity discriminator with an auxiliary task layer that aims at promoting the generation of domain-invariant features in the feature extractor and thus improving the performance of the model.
To show the effectiveness of the discriminator and the auxiliary task layer, we perform an ablation study and report the experimental results in Table \ref{Tab:abl_exp}.
For a fair comparison, we perform all experiments under the same hyper-parameter settings.
The baseline model shown in Table \ref{Tab:abl_exp} consists of only the feature extractor and the MRC-module and was trained on the labeled source domain dataset, SQuAD.
For the remaining two models, we use the labeled SQuAD and the unlabeled BioASQ training datasets simultaneously.
The addition of the discriminator enables the feature extractor in the baseline model to use the unlabeled BioASQ training datasets for generating domain-invariant feature representations.
This is achieved by using the dissimilarity measurements between the feature representations of the SQuAD and BioASQ training data.
As shown, after adding only the discriminator without the auxiliary task layer, the performance of the model improves from the baseline, suggesting the influence of the discriminator.
We explain this influence on the feature extractor more elaborately later in Section \ref{dom_ada}.
For the final experiment in the ablation study (Table \ref{Tab:abl_exp}), we use our whole model consisting of the domain similarity discriminator with the auxiliary task layer and notice an even further performance improvement.
The auxiliary task layer, in this study, constrains the changes in the task-relevant features in the domain similarity discriminator during training.
Thus, the improvement in model performance after incorporating the auxiliary task layer suggests that with the task layer, the domain similarity discriminator can better promote the generation of domain-invariant features that are simultaneously discriminative from the viewpoint of the MRC task in the source domain.
Moreover, as explained in Section \ref{sec:dom-sim-disc}, we further demonstrate the stabilizing capability of the auxiliary task layer in Section \ref{sec:res-stab}.

\subsection{Analysis}
In this section, we analyze different components of the proposed framework. We also study the domain adaptation capability and the strengths and weaknesses of the BioADAPT-MRC model.

\subsubsection{Triplet vs. distance-based loss}
\label{sec:res-loss}

\begin{table}[!htbp]
\centering
\caption{Average test scores with standard deviations (across three different seeds) for the contrastive experiments of discriminator loss functions – triplet loss and distance-based loss. The best scores are highlighted in bold.
\label{Tab:loss_compare}}
\resizebox{.8\textwidth}{!}{%
{\begin{tabular}{@{}l lll@{}}
\toprule
\multirow{2}{*}{Discriminator Loss}                       & \multicolumn{3}{c}{BioASQ-9b}\\
\cmidrule(lr){2-4}
                                             & SAcc     & LAcc     & MRR    \\
                                             \midrule
Distance-based & 0.5256$\pm$0.0202           & 0.7219$\pm$0.0104           & 0.6038$\pm$0.0105 \\
Triplet  & \textbf{0.5358$\pm$0.0029}  & \textbf{0.7321$\pm$0.0077}  & \textbf{0.6140$\pm$0.0035} \\
\hline
\end{tabular}}
}
\end{table}

\begin{figure}[!htpb]
\centerline{\includegraphics[width=1.\textwidth]{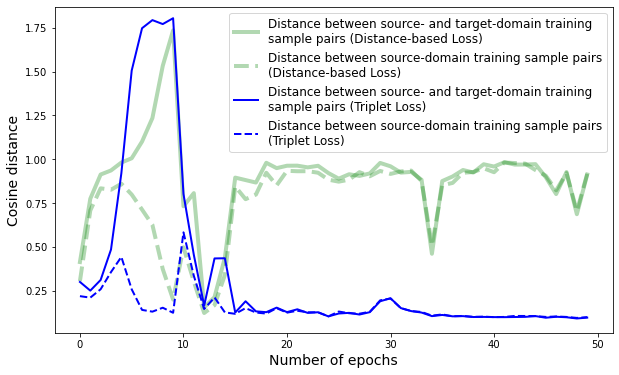}}
\caption{Per-epoch cosine distance between source-domain training sample pairs and source- and target-domain training sample pairs across 50 epochs.
}
\label{fig:cosdist_loss}
\end{figure}

BioADPT-MRC uses triplet loss to optimize the discriminator.
As explained in Section \ref{sec:dom-sim-disc}, we also consider using distance-based loss in place of triplet loss.
Table \ref{Tab:loss_compare} shows the results of the contrastive experiments of loss functions -- triplet loss and distance-based loss.
As shown in Table \ref{Tab:loss_compare}, the model with triplet loss outperforms the one with the distance-based loss with higher mean SAcc, LAcc, MRR, and lower standard deviation.

To further analyze this performance gap, we examine the trend of distance between source-domain training sample pairs and between source- and target-domain training sample pairs per epoch across 50 training epochs (Figure \ref{fig:cosdist_loss}).


Note that both experiments in Figure \ref{fig:cosdist_loss} are performed under the same seed.
As shown in Figure \ref{fig:cosdist_loss}, when we use distance-based loss, the cosine distance between either source-domain training sample pairs or source- and target-domain training sample pairs tends to be higher than when we use the triplet loss.
It may happen because in the adversarial framework, while distance-based loss focuses only on minimizing the distance between the source- and target-domain training samples without considering the distance between the source samples, triplet loss focuses on balancing both \cite{chen2017beyond, wang2017deep}.
As a result, triplet loss can minimize the domain shift to a greater extent than the distance-based loss and thus enable our framework to achieve higher performance.

\subsubsection{Stability analysis}
\label{sec:res-stab}

We examine the stability of BioADAPT-MRC and perform an error analysis of its performance by repeating the experiments for three different random seeds (10, 42, 2018).
Table \ref{Tab:stability} shows the test scores averaged across three seeds with standard deviations for BioASQ-7b, 8b, and 9b.

\begin{table}[!htbp]
\caption{Average test scores with standard deviations across experiments with three random seeds (10, 42, 2018) for initialization, to measure and compare the stability of BioADAPT-MRC. The best and the second-best scores are respectively highlighted in bold and italic.}
\vspace{-.2cm}
\label{Tab:stability}
\resizebox{1\textwidth}{!}{%
{\begin{tabular}{@{}l lll@{}}
\toprule
\multirow{2}{*}{Model}                       & \multicolumn{3}{c}{BioASQ-7b}\\
                                              \cmidrule(lr){2-4}
                                             & SAcc     & LAcc     & MRR  \\
                                             \toprule
AdaMRC       & \textit{0.4300$\pm$0.0029}  & 0.6152$\pm$0.0116           & \textit{0.5083$\pm$0.0076} \\
Baseline     & 0.4156$\pm$0.0127           & 0.6173$\pm$0.0101  & 0.5038$\pm$0.0033          \\
BioADAPT-MRC (no auxiliary layer) & 0.4177$\pm$0.0058 & \textit{0.6193$\pm$0.0105} & 0.5038$\pm$0.0076 \\
BioADAPT-MRC & \textbf{0.4465$\pm$0.0029}  & \textbf{0.6379$\pm$0.0029}  & \textbf{0.5237$\pm$0.0037} \\
\midrule
\multirow{2}{*}{Model}                       & \multicolumn{3}{c}{BioASQ-8b} \\
                                              \cmidrule(lr){2-4}
                                             & SAcc     & LAcc     & MRR \\
                                             \midrule
AdaMRC       & 0.3422$\pm$0.0083           & 0.5960$\pm$0.0143  & 0.4425$\pm$0.0031           \\
Baseline     & 0.3554$\pm$0.0125  & 0.5960$\pm$0.0162  & 0.4547$\pm$0.0152  \\
BioADAPT-MRC (no auxiliary layer) & \textit{0.3664$\pm$0.0113} & \textit{0.5982$\pm$0.0031} & \textit{0.4618$\pm$0.0080} \\
BioADAPT-MRC & \textbf{0.3797$\pm$0.0031}  & \textbf{0.6137$\pm$0.0083}  & \textbf{0.4750$\pm$0.0024}  \\
\midrule
\multirow{2}{*}{Model}                       & \multicolumn{3}{c}{BioASQ-9b}\\
                                              \cmidrule(lr){2-4}
                                             & SAcc     & LAcc     & MRR \\
                                             \midrule
AdaMRC       & 0.5276$\pm$0.0050  & 0.7239$\pm$0.0100  & 0.6068$\pm$0.0062 \\
Baseline     & 0.5174$\pm$0.0058           & 0.7198$\pm$0.0126           & 0.6018$\pm$0.0011          \\
BioADAPT-MRC (no auxiliary layer) & \textit{0.5337$\pm$0.0050} & \textit{0.7280$\pm$0.0153} & \textit{0.6127$\pm$0.0012} \\
BioADAPT-MRC & \textbf{0.5358$\pm$0.0029}  & \textbf{0.7321$\pm$0.0077}  & \textbf{0.6140$\pm$0.0035} \\
\hline
\end{tabular}}
}
\end{table}

We also report the averaged scores with standard deviations for our baseline, the BioADAPT-MRC model with no auxiliary layer, and a state-of-the-art method AdaMRC to compare model stability.
As shown in Table \ref{Tab:stability}, BioADAPT-MRC outperforms the other models with lower standard deviations, indicating higher stability of our framework.
Moreover, the scores from the BioADAPT-MRC models with and without the auxiliary layer indicate that the auxiliary task layer helps increase both performance and overall model stability.

\subsubsection{Masked vs. synthetic questions}
\label{sec:res-ques}

Recall that, BioADAPT-MRC uses a special token $[MASK]$ in place of the question tokens $Q$ for the unlabeled target-domain training samples.
The $[MASK]$ tokens are used to inform the encoder model that the question tokens are missing and maintain consistency in the structure of the tokenized samples.
Another approach to address the issue of missing questions in the target-domain training samples is to use synthetic questions \cite{wang-etal-2019-adversarial}.
In Table \ref{Tab:synth_ques}, we present the results of the contrastive experiments of these two approaches -- masked and synthetic questions.
Inspired by the success of the AdaMRC question-generator in various target domains such as news, Wikipedia, and web search log, we use it to generate the synthetic questions.


\begin{table}[!htbp]
\centering
\caption{Average test scores with standard deviations (across three different seeds) for experiments using synthetic questions and masked questions in the target-domain training dataset. The best and the second-best scores are respectively highlighted in bold and italic.}
\vspace{-.2cm}
\label{Tab:synth_ques}
\resizebox{.8\textwidth}{!}{%
{\begin{tabular}{@{}l lll@{}}
\toprule
\multirow{2}{*}{Questions}                       & \multicolumn{3}{c}{BioASQ-9b}\\ \cmidrule(lr){2-4}
                                             & SAcc     & LAcc     & MRR    \\
                                             \midrule
Synthetic  & \textit{0.5337$\pm$0.0087}           & \textit{0.7198$\pm$0.0126}           & \textit{0.6069$\pm$0.0059}          \\
Masked     & \textbf{0.5358$\pm$0.0029}  & \textbf{0.7321$\pm$0.0077}  & \textbf{0.6140$\pm$0.0035} \\
\hline
\end{tabular}}
}
\end{table}
\begin{figure}[htbp]
\centerline{\includegraphics[width=.8\textwidth]{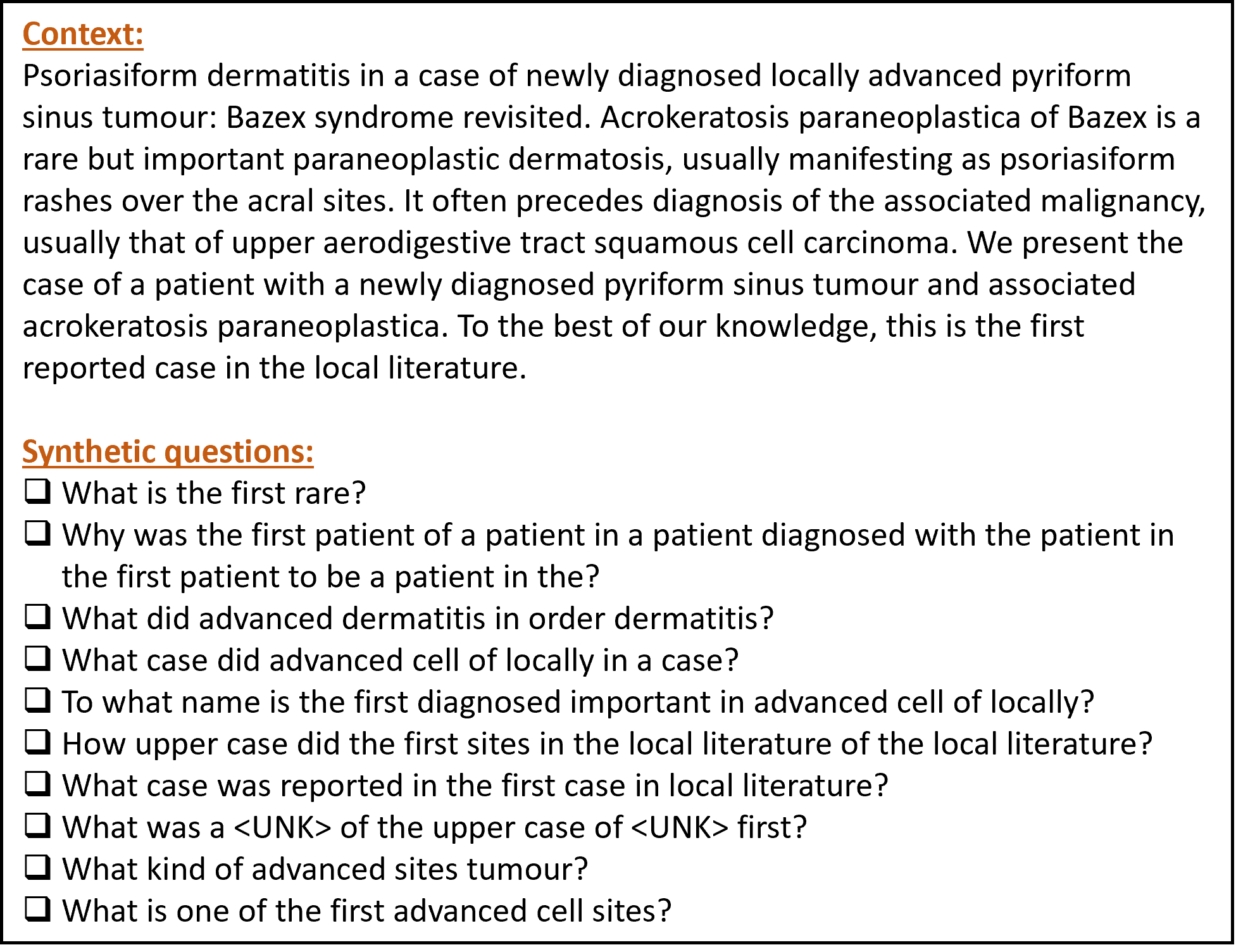}}
\caption{Synthetic questions generated from the first context of the BioASQ-9b training set using the \textit{question-generator} component of the AdaMRC model along with the context.
}\label{fig:synth}
\end{figure}

Table \ref{Tab:synth_ques} shows the average test scores for BioASQ-9b with standard deviations across three seeds.
We find that although the synthetic questions can noticeably improve performance over the baseline (see results for `Baseline' in Table \ref{Tab:stability}), BioADAPT-MRC with synthetic questions is unable to achieve better performance than with masked questions. 
It may happen because the biomedical domain differs from other domains such as news or Wikipedia in many linguistic dimensions such as syntax, lexicon, semantics \cite{lee2020biobert, verspoor2009textual}.
As a result, while the question-generator can generate meaningful questions for domains such as web, news, and movie reviews \cite{wang-etal-2019-adversarial}, it mostly generates incoherent questions for the biomedical domain (as shown in Figure \ref{fig:synth}), which can eventually hurt the performance of the model.

Moreover, synthetic-question generation also requires additional computational time -- generating questions from around 10,000 contexts using a trained question generator took approximately 3 hours with our computational resources.
Given the findings, we think that generating synthetic questions for the biomedical domain requires more attention and consider it as a future study.

\subsubsection{Semi-supervised setting}
\label{sec:res-semi}

As an additional experiment, we evaluate the BioADAPT-MRC framework under a semi-supervised setting where we combine labeled and unlabeled target-domain training data.
We perform four experiments where the ratios of labeled target-domain training samples are 0\%, 10\%, 50\%, and 80\% of the total target-domain training data.
Note that we choose the labeled target-domain data by random sampling.
Table \ref{Tab:labeled_9B} shows the test scores on BioASQ-9b when trained with varying ratios of labeled target-domain data.
As shown, with increased ratio of labeled samples in the target-domain training data, the performance scores also increase, which is expected.
These results suggest that our proposed framework is also effective in a semi-supervised setting.

\begin{table}[htbp]
\centering
\caption{Test scores for experiments using varied ratio of labeled target-domain training data. The best and the second-best scores are respectively highlighted in bold and italic.
\label{Tab:labeled_9B}}
\resizebox{.45\textwidth}{!}{%
{\begin{tabular}{@{}l lll@{}}\hline
\multirow{2}{*}{Ratio}                       & \multicolumn{3}{c}{BioASQ-9b}\\ \cmidrule(lr){2-4}
                                             & SAcc     & LAcc     & MRR    \\ \hline
0\%  & 0.5399                   & \textbf{0.7423}          & 0.6187                  \\
10\% & 0.5460                   & 0.7178                   & 0.6118                  \\
50\% & \textit{0.5644}          & \textit{0.7301}          & \textit{0.6264}         \\
80\% & \textbf{0.5828}          & \textit{0.7301}          & \textbf{0.6369}         \\
\hline
\end{tabular}}%
}
\end{table}



Note that although multiple labeled datasets in various sub-domains of biomedical MRC (such as scientific literature and clinical notes) have been made available in the past few years \cite{tsatsaronis2015overview, pampari-etal-2018-emrqa}, there is still a severe scarcity of labeled data in some other sub-domains that are linguistically different (e.g., consumer health biomedical-MRC) \cite{jin2022biomedical, nguyen2019question}.

\subsubsection{Results on emrQA}
\label{sec:res-emrqa}

We further validate our framework on another type of biomedical MRC dataset, emrQA \cite{pampari-etal-2018-emrqa}, built using unstructured textual electronic health records (EHRs) with questions reflecting the inquiries made by clinicians about patients' EHRs.
The dataset contains five subsets, three of which are extractive MRC datasets -- heart disease risk, relations, and medications.
For our experiments, we use the heart disease risk dataset as the target-domain dataset.
We randomly sample 10\% of the dataset as the test set.
To measure the performance, we use the widely used metrics for the extractive MRC task: Exact Match (EM) and F1-score \cite{baradaran2020survey}.
We compare the test scores with our baseline and the AdaMRC model.
Table \ref{Tab:emrqa} shows that BioADAPT-MRC improves the performance scores over baseline (9.67\% in EM and 10.86\% in F1) and AdaMRC (2.08\% in EM and 2.16\% in F1).
This experiment validates that BioADAPT-MRC can be applied to different types of biomedical-MRC datasets.

\begin{table}[!htbp]
\centering
\caption{Test scores for experiments using the emrQA (heart disease risk subset) data as the target-domain data. The best and the second-best scores are respectively highlighted in bold and italic.
\label{Tab:emrqa}}
\vspace{.2cm}
\resizebox{0.5\textwidth}{!}{%
{\begin{tabular}{@{}l lll@{}}\hline
\multirow{2}{*}{Model}                       & \multicolumn{2}{c}{emrQA}\\
                                              \cmidrule(lr){2-3}
                                & EM                      & F1                      \\ \hline
AdaMRC                          & \textit{0.1891}         & \textit{0.3815}         \\
Baseline                        & 0.1132                  & 0.2946                  \\
BioADAPT-MRC                    & \textbf{0.2099}         & \textbf{0.4031}         \\
\hline
\end{tabular}}%
}
\end{table}



We want to emphasize the fact that researchers have identified the unstructured clinical notes as inherently noisy and long with long-term textual dependencies \cite{cohen2013redundancy, pampari-etal-2018-emrqa, mahbub2022unstructured}.
We suspect that these phenomena may lead to an overall low EM and F1 score \cite{joshi-etal-2017-triviaqa}.
Hence, we think that achieving higher scores in an MRC task on EHRs requires additional and rigorous data pre-processing and leave it as a future work.

\subsubsection{Domain adaptation}
\label{dom_ada}
\begin{figure}[!htbp]
\centerline{\includegraphics[width=1\textwidth]{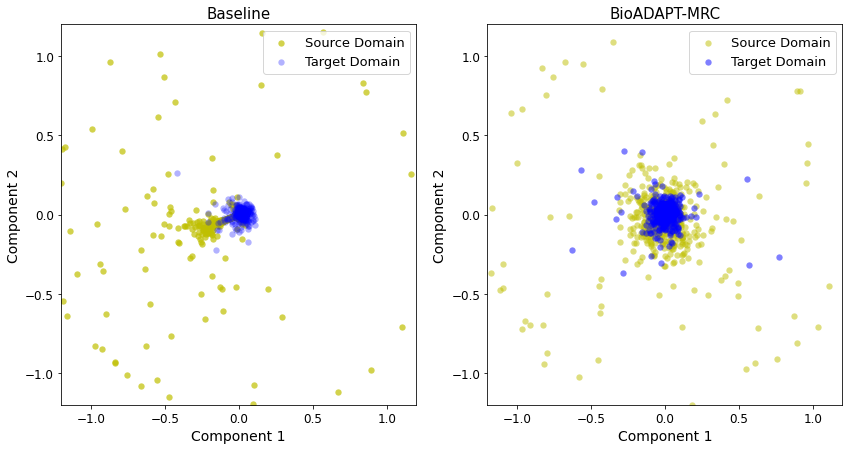}}
\caption{Multidimensional scaling (MDS) plot for the features (extracted from the baseline (left) and the BioADAPT-MRC (right) models) of all samples from the BioASQ-9b test set (blue) and a set of random samples from the SQuAD training set (yellow).
}
\label{fig:cluster}
\end{figure}

We show the influence of the domain similarity discriminator by plotting (Figure \ref{fig:cluster}) all samples from the BioASQ-9b test set and a set of random samples from the SQuAD training set.
We pick random samples from the SQuAD training set to match the number of samples in the BioASQ-9b test set.
As explained in Section \ref{sec:mat-fe}, we use the feature representation of the $[CLS]$ token as an accumulated representation of the whole input sequence.
Each feature representation of the $[CLS]$ token has a dimension of 768.
To reduce these dimensions into two for visualization, we use multidimensional scaling (MDS) \cite{kruskal1964multidimensional}.
We use MDS because it reduces the dimensions by preserving the dissimilarities between two data points in the original high-dimensional space.
Since we use cosine distance in the discriminator to measure the dissimilarity between two domains, as the dissimilarity measure in the MDS, we use the pairwise cosine distance.
The feature representations of the $[CLS]$ token on the left plot and the right plot in Figure \ref{fig:cluster} are generated by the feature extractors from the baseline model and the BioADAPT-MRC model, respectively.

For a fair comparison, the selection of random SQuAD training samples is the same for the baseline and BioADAPT-MRC models.
As shown, the features generated by the baseline model create two separate clusters for SQuAD and BioASQ-9b.
The features generated by the BioADAPT-MRC model, on the other hand, form two overlapping clusters implying the reduced dissimilarities between the source and target domains.
Interestingly, we notice that the data points from the BioASQ are closer to its cluster than those from the SQuAD.
It may be because, unlike SQuAD, the data in the BioASQ originate from one single domain, and thus the feature representations are more similar to one another.

\begin{figure}[!htpb]
\centerline{\includegraphics[width=1.\textwidth]{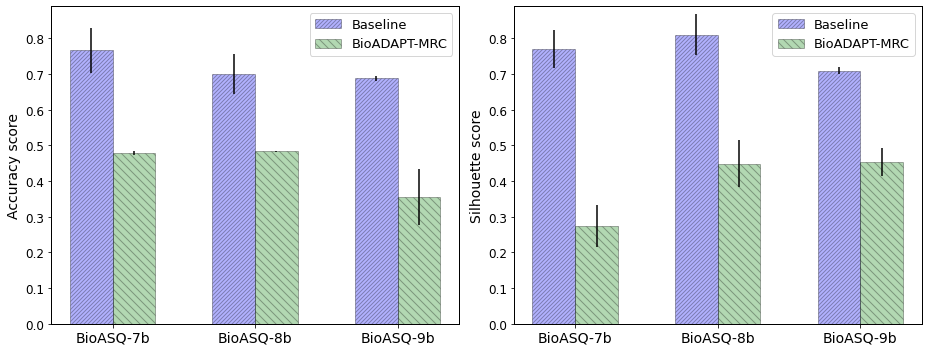}}
\caption{Mean accuracy scores (left) and mean silhouette scores (right) with standard deviations for DBSCAN clustering on BioASQ test sets and SQuAD.
}
\label{fig:cluster_acc_sil}
\end{figure}

To further analyze the quality of the clusters before and after introducing the domain similarity discriminator to the framework and thus to quantify the effect of domain adaptation, we perform DBSCAN clustering \cite{ester1996density}.
We perform clustering on the MDS components of the features for the $[CLS]$ tokens for the samples in the BioASQ test sets and the random samples from the SQuAD training set.
Considering the bias of random sampling, for each BioASQ test set, we select five sets of random samples from the SQuAD training set and report the mean accuracy and silhouette scores with standard deviation in Table \ref{Tab:cluster_perf} and Figure \ref{fig:cluster_acc_sil}.
We use the DBSCAN clustering because it views clusters as high-density regions where the distance between the samples is measured by a distance metric, providing flexibility in shapes and numbers of clusters.
We describe the selected hyperparameters for the DBSCAN algorithm implementation below:

As the distance measure for DBSCAN, we choose pairwise cosine distance.
We implement the DBSCAN algorithm using the scikit-learn tool \cite{pedregosa2011scikit} and tune the two essential hyperparameters: $eps$ and $min\_samples$.
The details on these two hyperparameters can be found on the scikit-learn documentation.
After optimization, we select 0.005, 0.01, and 0.001 as the $eps$ for BioASQ-7b, 8b, and 9b, respectively, and 20 as the $min\_samples$ for all three test sets.
Note that, for a fair comparison, for each test set, the choice of hyperparameters are the same for the baseline and BioADAPT-MRC models.

Table \ref{Tab:cluster_perf} and Figure \ref{fig:cluster_acc_sil} shows that DBSCAN can identify two clusters with high accuracy when the features of the samples are extracted from the baseline model.
The accuracy goes down when the features of the same samples are extracted from the BioADAPT-MRC model as they form a single cluster.
Moreover, we analyze the silhouette scores to understand the separation distance between clusters.
\begin{table}[!htbp]
\caption{Mean accuracy and mean silhouette scores with standard deviations for DBSCAN clustering on the MDS components.
The MDS dimensionality reduction technique was applied to reduce the dimension of the $[CLS]$ token representations for the samples in the BioASQ test sets and a set of random samples from the SQuAD training set.
A $[CLS]$ token represents the accumulated representations a sample in the dataset.
}
\label{Tab:cluster_perf}
\resizebox{1\textwidth}{!}{%
{\begin{tabular}{@{}l ll ll@{}}\hline
                              & \multicolumn{2}{c}{Mean accuracy ($\pm$standard deviation)} & \multicolumn{2}{c}{Mean silhouette score ($\pm$standard deviation)}  \\
                                \cmidrule(lr){2-3}                                               \cmidrule(lr){4-5}
                              & Baseline                            & BioADAPT-MRC                            & Baseline                                & BioADAPT-MRC             \\ \hline
BioASQ-7b                     & 0.7664($\pm$0.0625)                 & 0.4786($\pm$0.0051)                & 0.7714($\pm$0.0535)                     & 0.2745($\pm$0.0585) \\
BioASQ-8b                     & 0.7011($\pm$0.0563)                 & 0.4838($\pm$0.0015)                & 0.8111($\pm$0.0577)                     & 0.4484($\pm$0.0660) \\
BioASQ-9b                     & 0.6880($\pm$0.0075)                 & 0.3561($\pm$0.0782)                & 0.7100($\pm$0.0089)                     & 0.4540($\pm$0.0398) \\ \hline                 
\end{tabular}}%
}
\end{table}

The range of silhouette score is $[-1,1]$.
A score of 1 indicates that the clusters are highly dense and clearly distinguishable from each other whereas -1 refers to incorrect clustering.
A score of zero or near zero indicates indistinguishable or overlapping clusters.
As shown in Table \ref{Tab:cluster_perf} and Figure \ref{fig:cluster_acc_sil}, in this case, the high silhouette scores (closer to 1) for the baseline model reflect that the feature representations of the samples from the same domain are very similar to its own cluster compared to the other one.
On contrary, the low silhouette scores (closer to zero) for the BioADAPT-MRC model indicate that the feature representations of the samples from both domains are very similar to one another.
These results show the effectiveness of the domain similarity discriminator in the BioADAPT-MRC framework.

\subsubsection{Motivating example}

\begin{figure*}[!htbp]
\centerline{\includegraphics[width=1.\textwidth]{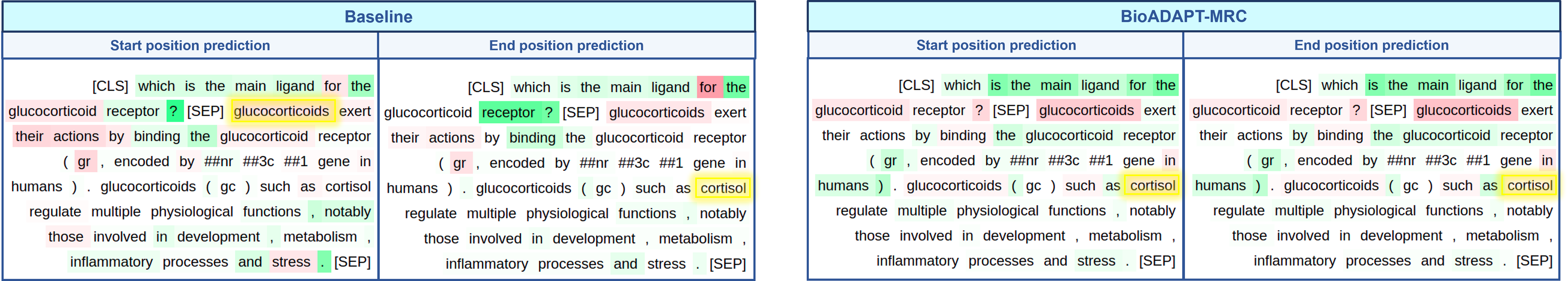}}
\caption{Motivating example from the BioASQ-9b test set.
The left and right plots show the word importance for answer span prediction given by the baseline and BioADAPT-MRC models.
The green and red colors show positive and negative word importance, respectively.
The density of color shows the amount of importance.
The highlighted word in yellow represents either the predicted start or end tokens.}\label{fig:motivating_ex}
\end{figure*}

Considering the variability of the predicted answers in an MRC-task, we present a motivating example to demonstrate how the word importance may impact the answer predictions and thus the performance of the biomedical-MRC task. The example in Figure \ref{fig:motivating_ex} shows the effectiveness of BioADAPT-MRC over the baseline model for the given sample.


This example is selected randomly from the BioASQ-9b test set from the samples incorrectly predicted by the baseline model and correctly predicted by the BioADAPT-MRC model.
The question is separated from the context by the $[SEP]$ token and the correct answer, in this case, is $cortisol$.
The predicted start and end tokens are highlighted in yellow.
The positive and negative importance of the words are highlighted  in green and red, respectively.
The intensity of the color shows the amount of importance.
Higher intensity reflects higher importance.
We calculate the word importance using the \textit{Integrated Gradients} algorithm \cite{sundararajan2017axiomatic} implemented in the Captum tool \cite{kokhlikyan2020captum}.
Integrated gradients measure the importance of the word (in predicting the answer span) by computing the gradient of the predicted output with respect to the specified input word \cite{sundararajan2017axiomatic}.

In this particular example, the start and end token being the same, we expect that the correct prediction of the MRC model should provide similar importance to the words for the start and end position predictions.
However, while this is true for the BioADAPT-MRC model (right plot), for the baseline (left plot) we notice a different pattern in word importance while predicting for start and end positions and thus result in wrong answer span prediction.
It shows the effectiveness of our model over the baseline for this example.

\subsubsection{Error analysis}
We analyze the strengths and weaknesses of our approach by performing two-fold error analysis.
We focus on two aspects of the MRC dataset -- types of questions and answers.
Through this error analysis, we aim to answer the following questions:
(i) What types of questions can be answered after domain adaptation that could not be answered before?
(ii) What types of answers can be identified by the proposed approach after domain adaptation?
(iii) What types of items does the model struggle with, even with the domain adaptation component?

\begin{figure}[!htbp]
\centerline{\includegraphics[width=1.\textwidth]{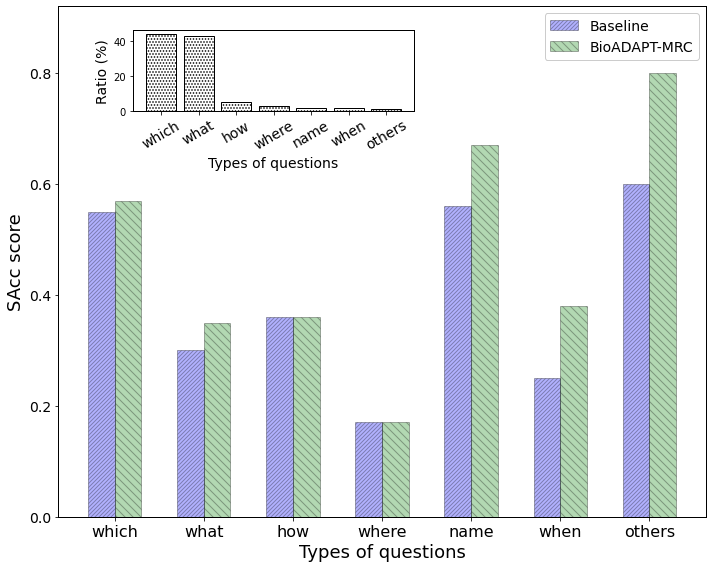}}
\caption{
Error analysis of BioADAPT-MRC, in comparison with the baseline model, depending on the types of questions in the BioASQ test sets.
}
\label{fig:qtype}
\end{figure}
\begin{figure}[!htbp]
\centerline{\includegraphics[width=1.\textwidth]{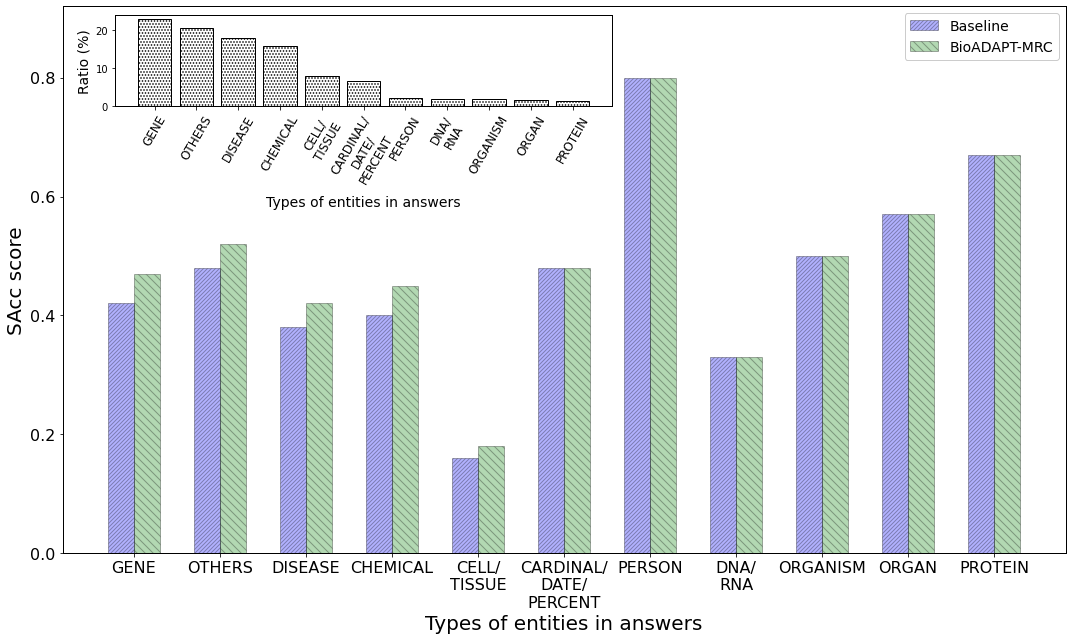}}
\caption{
Error analysis of BioADAPT-MRC, in comparison with the baseline model, depending on the types of answers in the BioASQ test sets.
}
\label{fig:ner}
\end{figure}

\begin{figure}[!htbp]
\centerline{\includegraphics[width=1\textwidth]{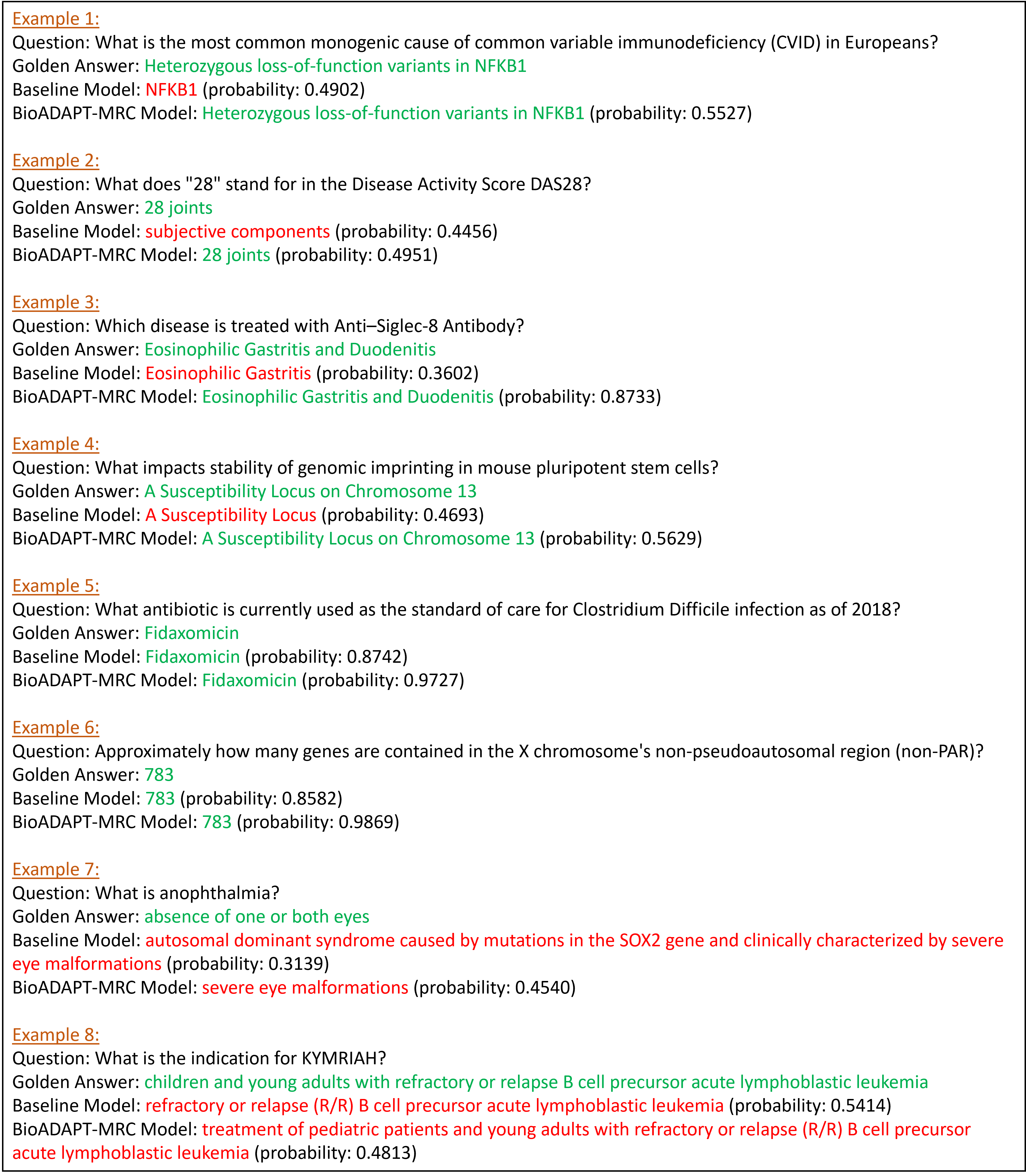}}
\caption{
Example question-answer pairs from the test sets demonstrating the strengths
and weaknesses of the BioADAPT-MRC model over the baseline
model. 
The green and red colors show correctly and incorrectly predicted answers, respectively.
}
\label{fig:example_qa_ano}
\end{figure}

We use the SAcc score to perform the error analysis since the ultimate goal for any MRC system is to predict correct answer spans with the highest probability, reflected in the SAcc score.
We categorize the test sets based on the following types of questions: \textit{which, what, how, where, when}, and \textit{name} (an example question for \textit{name} -- ``Name a CFL2 mutation which is associated with nemaline myopathy?'').
We find that the most prevalent question types in the test sets are \textit{what} (43\%) and \textit{which} (44\%) (embedded in Figure \ref{fig:qtype}).
We also find that after adding the domain adaptation module, the SAcc score increases by 52\%, 20\%, 17\%, and 4\% for question types -- \textit{when, name, what,} and \textit{which}, respectively (Figure \ref{fig:qtype}).
It indicates that the domain adaptation module can help increase the model's capability to answer these types of questions with higher probability.
For question types \textit{how} and \textit{where}, we do not notice further improvement, even with the domain adaptation module.

To analyze the types of answers that can be identified after introducing the domain adaptation module, we categorize the test sets based on the named entities in the answers.
We use named entity recognition (NER) algorithms from the widely used spaCy library \cite{honnibal2017spacy}.
Figure \ref{fig:ner} shows that the answers in the test sets mainly consists of entities such as \textit{GENE (23\%), DISEASE (18\%), CHEMICAL (16\%), CELL/TISSUE (8\%),} and \textit{CARDINAL/DATE/PERCENT (7\%)}.

As shown in Figure \ref{fig:ner}, after adding the domain adaptation module, the model has been able to identify the \textit{CHEMICAL, CELL/TISSUE} entities with the highest SAcc scores (improvement by 13\% from the baseline).
We also notice an improvement in the SAcc scores for \textit{GENE} and \textit{DISEASE} entities.
However, we notice no improvement in the SAcc scores for \textit{CARDINAL/DATE/PERCENT, PERSON, DNA/RNA, ORGANISM, ORGAN, and PROTEIN} entities even after adding the domain adaptation module.
Given the results, we would like to emphasize that in both of the aforementioned analyses, for some categories, we do not have a large enough sample set to draw a definite conclusion about the effectiveness of the domain adaptation module and hence requires further future investigation.

To provide further evidence about what the model learned, we present eight example question-answer pairs demonstrating the strengths and weaknesses of the BioADAPT-MRC model over the baseline model (Figure \ref{fig:example_qa_ano}).
We randomly select two examples from each of the following four categories:
(i) mispredicted by baseline, correctly predicted by BioADAPT-MRC,
(ii) incompletely predicted by baseline, correctly predicted by BioADAPT-MRC,
(iii) correctly predicted by both baseline and BioADAPT-MRC, and
(iv) mispredicted by both baseline and BioADAPT-MRC.
In examples 1-8, the answer spans respectively contain GENE, CARDINAL, DISEASE, CELL, CHEMICAL, CARDINAL, ORGAN, and DISEASE entities.
As shown in examples 1-6, BioADAPT-MRC model can identify the answer span better than the baseline model with higher probability score.
However, the probability scores (i.e., the prediction capability) can be further improved.
Moreover, examples 7 and 8 provide additional motivation for future investigation of the reason behind the misprediction of the model.

The results from the overall error analysis indicate that while the BioADAPT-MRC model does well under various scenarios, there is still significant room for potential improvement.


\section{Conclusion}

Biomedical machine reading comprehension is a crucial and emerging task in the biomedical domain pertaining to natural language processing.
Biomedical-MRC aims at perceiving complex contexts from the biomedical domain and helping medical professionals to extract information from them.
Most MRC methods rely on a high volume of human-annotated data for near or similar to human-level performance.
However, acquiring a labeled MRC dataset in the biomedical domain is expensive in terms of domain expertise, time, and effort, creating the need for transfer learning from a source domain to a target domain.
Due to variance between two domains, directly transferring an MRC model to the target domain often negatively affects its performance.
We propose a framework for biomedical machine reading comprehension, BioADAPT-MRC, addressing the issue of domain shift by using a domain adaptation technique in an adversarial learning setting.
We use a labeled MRC dataset from a general-purpose domain (source domain) along with unlabeled contexts from the biomedical domain (target domain) as our training data.
We introduce a domain similarity discriminator, aiming to reduce the domain shift between the general-purpose domain and biomedical domain to help boost the performance of the biomedical-MRC model.
We validate our proposed framework on three widely used benchmark datasets from the biomedical question answering and semantic indexing challenge, BioASQ.
We comprehensively demonstrate that without any label information in the target domain during training, the BioADAPT-MRC framework can achieve state-of-the-art performance on these datasets.
We perform an extensive quantitative study on the domain adaptation capability using dimensionality reduction and clustering techniques and show that our framework can learn domain-invariant feature representations.
Additionally, we extend our framework to a semi-supervised setting and demonstrate that our framework can be efficiently applied even with varying ratios of labeled data. We perform a two-fold error analysis to investigate the shortcomings of our framework and provide motivation for further future investigation and improvement.

We conclude that BioADAPT-MRC may be beneficial in healthcare systems as a tool to efficiently retrieve information from complex narratives and thus save valuable time and effort of the healthcare professionals.

The following are some future research directions that can originate from this work:
(i) Developing a synthetic question-answer generator specializing in the biomedical domain.
(ii) Focusing on rigorous data pre-processing for the MRC task on unstructured clinical notes.
(iii) Performing further investigation on the cases where BioADAPT-MRC struggles to improve over the baseline model.
(iv) Applying BioADPT-MRC to other NLP applications in the biomedical domain that suffer from labeled-data-scarcity issues.
Such applications are biomedical named entity recognition, clinical negation detection, etc.
(v) Analyzing the robustness of the domain-invariant feature representations learned by the BioADAPT-MRC model against meticulously crafted adversarial attack scenarios that may leverage syntactic and lexical knowledge-base from the dataset.

\begin{table}[!htbp]
    \centering
    \caption{Scores from BioASQ-9b leaderboard.
    Here, `Average' implies to the average of mean SAcc, mean LAcc, and mean MRR scores}
    \label{tab:9b_lead}
    \begin{tabular}{@{}l llll@{}}
    \toprule
                  System &  Mean SAcc &  Mean LAcc &  Mean MRR &  Average \\
    \midrule
                 Ir\_sys2 &   0.503065 &   0.662573 &  0.566660 &          0.577433 \\
                  lalala &   0.466259 &   0.668706 &  0.544162 &          0.559709 \\
                 Ir\_sys1 &   0.478536 &   0.638055 &  0.544579 &          0.553723 \\
                 Ir\_sys4 &   0.447864 &   0.644181 &  0.518606 &          0.536884 \\
          LASIGE\_ULISBOA &   0.411042 &   0.650313 &  0.505514 &          0.522290 \\
                 Ir\_sys3 &   0.423315 &   0.631908 &  0.497349 &          0.517524 \\
        LASIGE\_ULISBOA\_3 &   0.404896 &   0.631904 &  0.497517 &          0.511439 \\
        LASIGE\_ULISBOA\_2 &   0.404911 &   0.631901 &  0.482426 &          0.506413 \\
               KU-DMIS-5 &   0.404922 &   0.607362 &  0.480883 &          0.497722 \\
               KU-DMIS-2 &   0.404922 &   0.601233 &  0.480654 &          0.495603 \\
      bio-answerfinder-2 &   0.453973 &   0.539867 &  0.492339 &          0.495393 \\
        bio-answerfinder &   0.453973 &   0.539867 &  0.492339 &          0.495393 \\
               KU-DMIS-3 &   0.404902 &   0.595119 &  0.481271 &          0.493764 \\
               KU-DMIS-1 &   0.398782 &   0.582811 &  0.469935 &          0.483843 \\
               KU-DMIS-4 &   0.386517 &   0.588975 &  0.463591 &          0.479695 \\
            Best factoid &   0.368108 &   0.527612 &  0.430671 &          0.442130 \\
            AUEB-System4 &   0.368087 &   0.515336 &  0.426766 &          0.436729 \\
              Best yesno &   0.355813 &   0.503080 &  0.414328 &          0.424407 \\
            AUEB-System3 &   0.349711 &   0.484645 &  0.408483 &          0.414280 \\
            AUEB-System2 &   0.368114 &   0.453992 &  0.407663 &          0.409923 \\
            UoT\_baseline &   0.337436 &   0.484642 &  0.397855 &          0.406644 \\
     UoT\_multitask\_learn &   0.337421 &   0.478522 &  0.392840 &          0.402928 \\
            AUEB-System1 &   0.318998 &   0.429440 &  0.365547 &          0.371328 \\
        UoT\_allquestions &   0.214723 &   0.380357 &  0.276910 &          0.290663 \\
         BioASQ\_Baseline &   0.079759 &   0.325160 &  0.167685 &          0.190868 \\
    \bottomrule
    \end{tabular}
\end{table}


\section*{Author contributions}

Maria Mahbub: Conceptualization, Data curation, Formal analysis, Investigation, Methodology, Software, Validation, Visualization, Writing – original draft, Writing – review \& editing.
Sudarshan Srinivasan: Writing – review \& editing.
Edmon Begoli: Supervision, Writing – review \& editing.
Gregory D. Peterson: Supervision, Writing – review \& editing.

\section*{Funding}
This work was supported by Department of Veterans Affairs, VHA Office of Mental Health and Suicide Prevention.
This manuscript has been authored by UT-Battelle, LLC, under contract DE-AC05-00OR22725 with the US Department of Energy (DOE).
The US government retains and the publisher, by accepting the article for publication, acknowledges that the US government retains a nonexclusive, paid-up, irrevocable, worldwide license to publish or reproduce the published form of this manuscript, or allow others to do so, for US government purposes.
DOE will provide public access to these results of federally sponsored research in accordance with the DOE Public Access Plan (\url{http://energy.gov/downloads/doe-public-access-plan}).

This research used resources of the Knowledge Discovery Infrastructure at the Oak Ridge National Laboratory, which is supported by the Office of Science of the U.S. Department of Energy under Contract No. DE-AC05-00OR22725 and the Department of Veterans Affairs Office of Information Technology Inter-Agency Agreement with the Department of Energy under IAA No. VA118-16-M-1062.

\textbf{Disclaimer:} The views and opinions expressed in this manuscript are those of the authors and do not necessarily represent those of the Department of Veterans Affairs, Department of Energy, or the United States Government. The funders had no role in study design, data collection and analysis, decision to publish, or preparation of the manuscript.




\bibliographystyle{plain}

\bibliography{bibfile}

\end{document}